\documentclass[10pt,twocolumn,letterpaper]{article}

\usepackage[pagenumbers]{cvpr}      % To produce the REVIEW version
\usepackage{CJKutf8}

\definecolor{cvprblue}{rgb}{0.21,0.49,0.74}
\usepackage[pagebackref,breaklinks,colorlinks,allcolors=cvprblue]{hyperref}
\usepackage{bbding}
\usepackage{multirow} 

\def\paperID{*****} % *** Enter the Paper ID here
\def\confName{CVPR}
\def\confYear{2026}

\title{CineScale: Tuning-Free High-Resolution Video Generation}

\author{
Gordon Chen$^{1,*}$ \quad
Haonan Qiu$^{1,*}$ \quad
Ning Yu$^{2,\dagger}$ \quad
Ziqi Huang$^{1}$ \quad
Paul Debevec$^{2}$ \quad
Ziwei Liu$^{1,\dagger}$
\\[2pt]
$^{1}$Nanyang Technological University
\quad
$^{2}$Netflix Eyeline Studios
\\
$^{*}$Equal contribution
\quad
$^{\dagger}$Corresponding authors
}

\begin{document}

\twocolumn[{%
            \renewcommand\twocolumn[1][]{#1}%
            \maketitle

            \vspace{-25pt}
            \begin{center}
                \textbf{Project Page:}
                \url{https://eyeline-labs.github.io/CineScale/}
            \end{center}
            \vspace{5pt}

            \begin{center}
                \centering
                \includegraphics[width=0.99\textwidth]{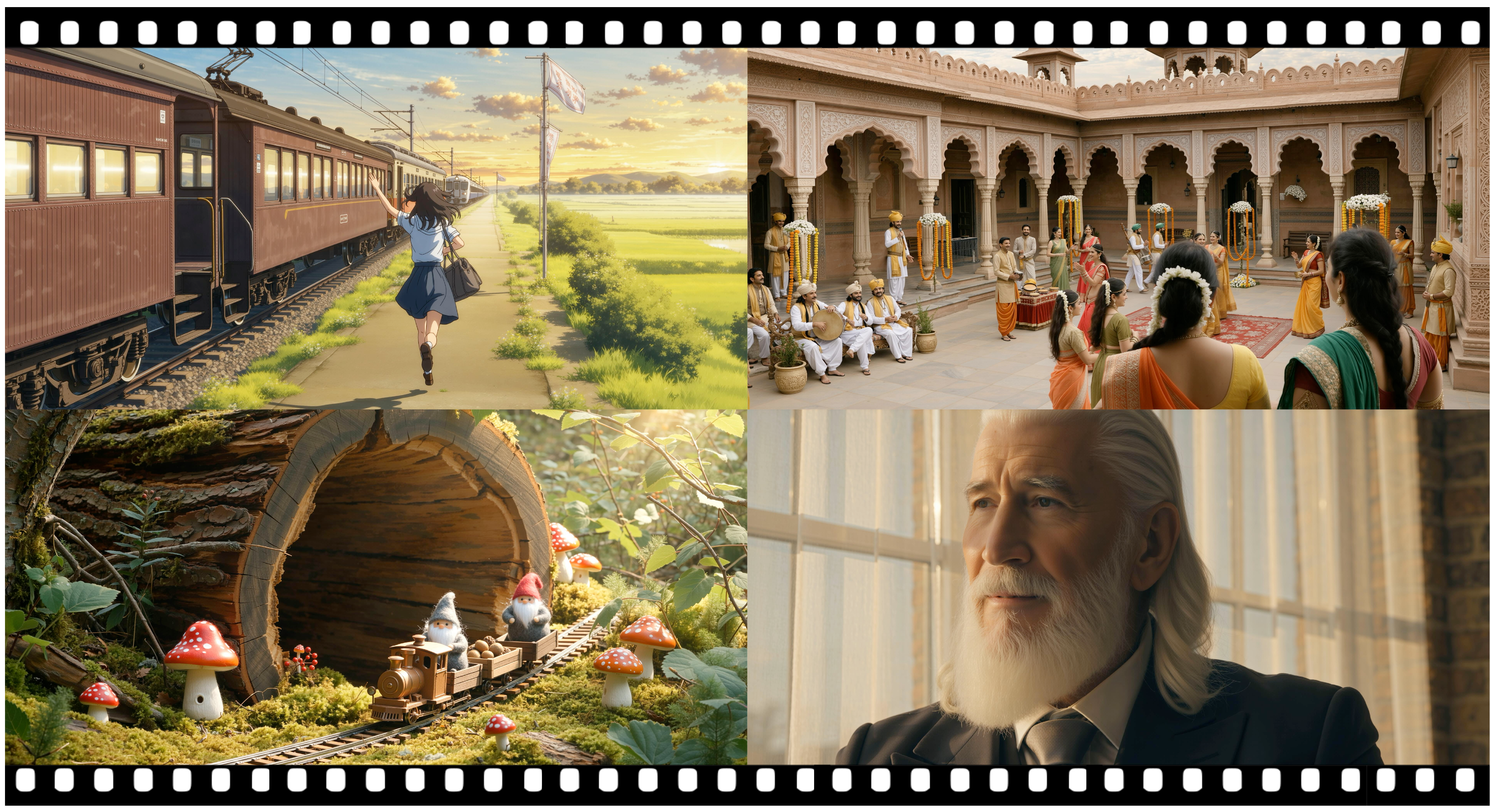}
                \captionof{figure}{\textbf{CineScale} enables pretrained diffusion models, including Wan2.2, Hunyuan, and CogVideoX, to generate high-fidelity 4K videos without fine-tuning, even when trained only on video data up to 720p. Watch our \href{https://www.youtube.com/watch?v=WMK8aZtKAec}{demo video}.}
                \label{teaser}
            \end{center}
        }]

\begin{abstract}
Video diffusion models have achieved remarkable progress in recent years, yet generating high-resolution videos remain a fundamental challenge. Most video generators are trained at limited spatial resolutions due to the scarcity of high-resolution 4K video data and the prohibitive computational cost of large-scale training on such data. Most video diffusion models are trained on 720p videos and are therefore effectively limited to generating videos at similar resolutions during inference. To address this gap, we propose \textit{CineScale}. CineScale, to the best of our knowledge, is the first tuning-free inference framework enabling pretrained video diffusion models to generate high-quality videos at resolutions far beyond those seen during training. Our key observation is that generation quality degrades at higher resolutions because positional encodings shift beyond their training distribution, producing blurred details and structurally incoherent videos. To address this gap, we introduce Adaptively Rectified RoPE. Our extensive experiments show that CineScale enables pretrained diffusion models, despite never being trained on high-resolution data, to generate high-fidelity 4K video without any fine-tuning, improving local detail and sharpness while preserving temporal coherence. This demonstrates that high-resolution generation capabilities can be unlocked purely at inference time.
\end{abstract}

\section{Introduction}
\label{sec:introduction}

Diffusion models have revolutionized visual generation~\cite{sdxl,pixart-alpha,wang2023modelscope,chen2024videocrafter2,yang2024cogvideox,zhang2023adding}, enabling anyone to generate stunning videos from text prompts. However, the limited availability of high-resolution 4K or 8K visual data and the substantially greater computational costs required to model such data mean that current diffusion models are typically trained at extremely low spatial resolutions. For example, SD 1.5~\cite{ldm} and SDXL~\cite{sdxl} are trained at image resolutions of $512^2$ and $1024^2$, respectively; VideoCrafter2~\cite{chen2024videocrafter2} is trained on $512\times320$ videos; and recent video diffusion models, including Wan~\cite{wan2025} and HunyuanVideo~\cite{kong2024hunyuanvideo}, are trained at resolutions up to 720p. Consequently, these models struggle to generate high-quality videos with fine-grained detail beyond their native training resolutions. This raises an important question: Can we adapt existing video models, typically trained at 720p, to generate videos with 4K fidelity and 4K resolution without requiring additional supervision on high-resolution data?

\begin{figure}[t]
\centering
\includegraphics[width=0.99\linewidth]{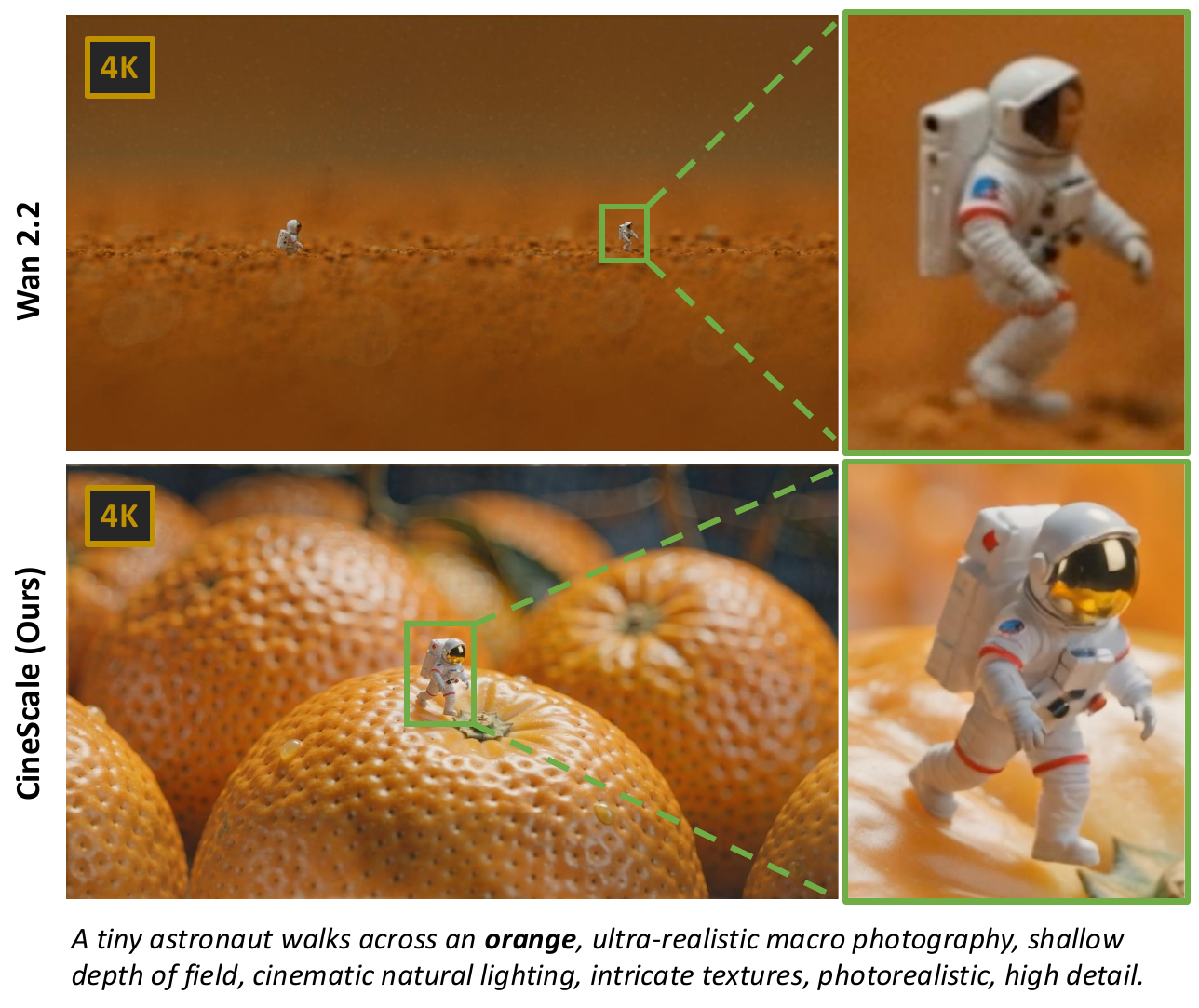}
\caption{\textbf{DiT Scaling Problem.} State-of-the-art DiT diffusion models, such as Wan2.2, struggle to generate videos at resolutions substantially higher than their native training resolution (up to 720p), resulting in blurring, degraded text-video alignment, and poor global coherence, which we show in this paper is mainly caused by out-of-distribution position encodings.
}
\end{figure}

\begin{figure}[t]
\centering
\includegraphics[width=0.99\linewidth]{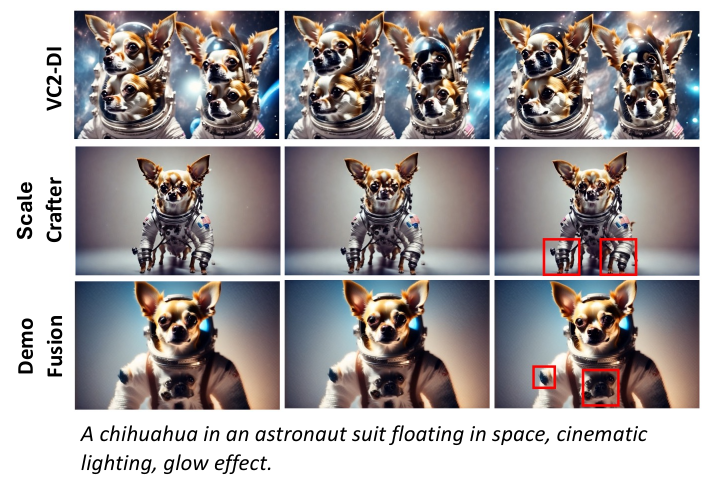}
\caption{\textbf{Existing U-Net-based approaches to high-resolution image generation} do not readily extend to video generation and exhibit static-ness and global incoherence.  We used VideoCrafter2~\cite{chen2024videocrafter2} as the base U-Net model.
}
\label{fig:existing_methods}
\end{figure}

Recent tuning-free methods have shown that pretrained U-Net image diffusion models can generate images at resolutions beyond those encountered during training. ScaleCrafter~\cite{he2023scalecrafter} enlarges convolutional receptive fields, DemoFusion~\cite{du2024demofusion} combines local patch denoising with global image context, and FouriScale~\cite{huang2024fouriscale} enforces global coherance through frequency-domain filtering. However, as demonstrated in Fig.~\ref{fig:existing_methods} and ~\ref{fig:gap}, these methods are designed for U-Net image models and do not readily extend to DiT-based video models. As a result, directly extending them to video generation leads to blur, temporal inconsistency, or unstable structure. To bridge this gap, recent methods rely on extensive fine-tuning with 4K video data~\cite{ultravideo,hu2026ultragen, zhao2026luve}, incurring substantial computational and data costs. To the best of our knowledge, no existing method enables video diffusion models trained at low resolutions to generate substantially higher-resolution, high-fidelity videos at inference time without additional supervision.

To address this gap, we propose CineScale. CineScale is a novel tuning-free, plug-and-play framework enabling DiT video diffusion models trained on low resolution videos to unlock native 4K resolution generation capabilities. In this paper, we observe that applying DiT models at resolutions far beyond those encountered during training causes their positional encodings to operate out of distribution, degrading both local sharpness and global structural coherence. Interestingly, these challenges closely resemble those encountered in extending the context length of large language models, where enlarged token sequences similarly lead to positional encoding extrapolation~\cite{chen2023extending,peng2023yarn}. Building on these insights, we propose a novel inference-time positional encoding scheme for DiT video diffusion models: \textit{Adaptively Rectified RoPE}, which corrects the out-of-distribution drift of positional encodings at inference time without any additional training. CineScale extends the native resolution capability of state-of-the-art open-source video diffusion models across both text-to-video (T2V) and image-to-video (I2V). Our contributions are summarized as follows:
\begin{itemize}

\item We propose CineScale, a tuning-free, plug-and-play framework enabling DiT video diffusion models trained on low resolution videos to unlock native 4K resolution generation capabilities at inference-time.
\item We identify out-of-distribution spatial RoPE offsets as a major source of visual degradation and structural incoherence when DiT-based diffusion models generate videos at high resolutions, and address this issue with Adaptively Rectified RoPE.
\item We empirically evaluate our approach across multiple metrics, demonstrating consistent gains over existing tuning-free high-resolution baselines.
\end{itemize}

\section{Related Work}
\label{sec:related}

\noindent\textbf{Diffusion Models for Video Generation.} Latent video diffusion models have significantly advanced text-to-video generation.
LVDM~\cite{lvdm} introduced hierarchical latent video diffusion for long video synthesis, while VideoCrafter2~\cite{chen2023videocrafter1,chen2024videocrafter2} and SVD~\cite{svd} demonstrated high-quality large-scale latent video generation. More recently, transformer-based diffusion architectures have become the dominant paradigm for video generation. Representative models include CogVideoX~\cite{yang2024cogvideox}, Pyramid Flow~\cite{jin2024pyramidal}, Mochi~\cite{genmo2024mochi}, LTX~\cite{HaCohen2024LTXVideo}, HunyuanVideo~\cite{kong2024hunyuanvideo}, and Wan~\cite{wan2025}, which consistently outperform earlier U-Net-based approaches in both generation quality and scalability.
Our method builds on Wan~\cite{wan2025} but is generally applicable to video diffusion transformers employing spatial RoPE.

\noindent\textbf{Higher-Resolution Visual Generation.} Training-free high-resolution generation methods improve inference without additional optimization or high-resolution supervision. MultiDiffusion~\cite{bar2023multidiffusion} applies a pretrained denoiser independently to overlapping latent patches and fuses their predictions. DemoFusion~\cite{du2024demofusion} extends this patch-wise formulation for SDXL by combining progressive upscaling with skip-residual guidance and dilated sampling, while ElasticDiffusion~\cite{haji2024elasticdiffusion} decomposes the U-Net denoising signal into patch-derived local detail and reference-derived global structure. Other methods modify U-Net components more directly. ScaleCrafter~\cite{he2023scalecrafter} dilates convolutional kernels to enlarge the U-Net's receptive field, whereas HiDiffusion~\cite{zhang2024hidiffusion} introduces a resolution-aware U-Net and replaces its global self-attention with modified shifted-window attention. FreeScale~\cite{qiu2025freescale} combines self-cascade upscaling with restrained dilated convolutions and frequency-selective fusion of global and local attention features.

\begin{figure}[t]
\centering
\includegraphics[width=0.99\linewidth]{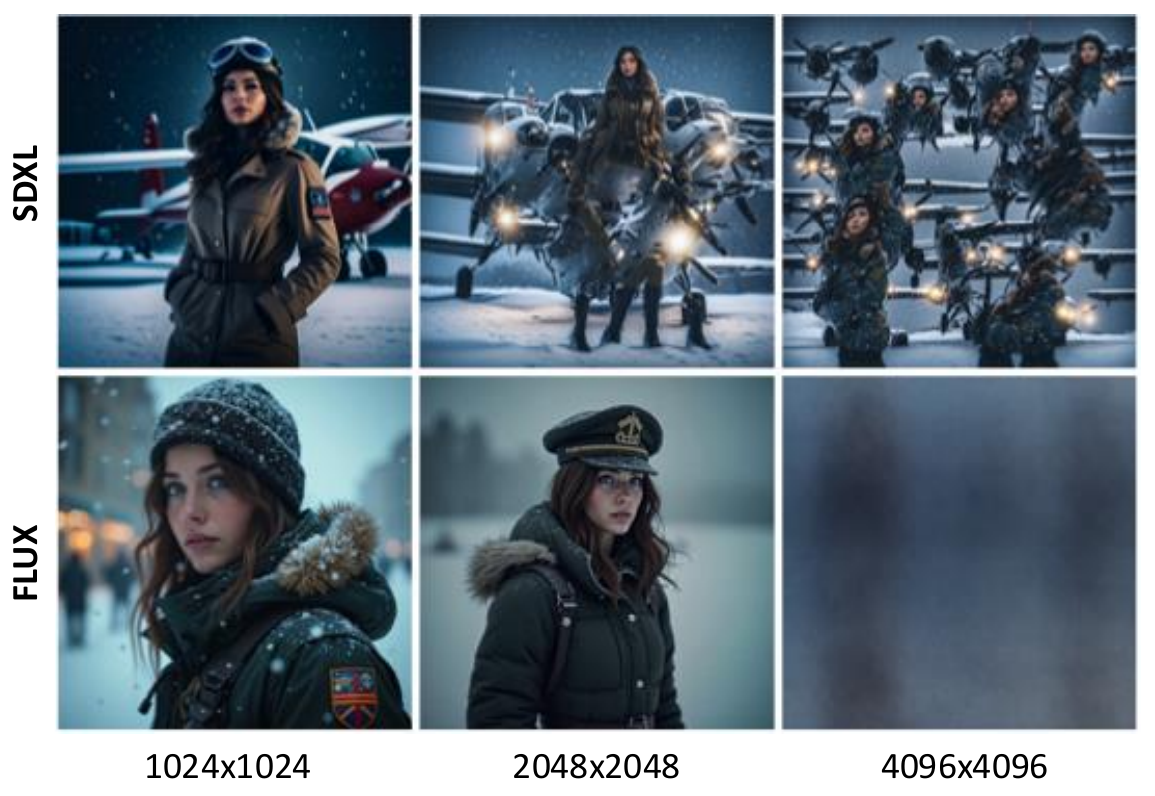}
\caption{\textbf{Distinct Failure Modes of U-Net and DiT Diffusion Models.} Diffusion models built on U-Nets suffer from low receptive field resulting in global incoherence while those built on DiTs suffer out-of-distribution position encodings resulting in blurring.
}
\label{fig:gap}
\end{figure}

Despite their diversity, these methods are primarily designed for image diffusion and do not readily extend to video DiTs, where the observed issue is that spatial RoPE coordinates fall outside the positional-encoding distribution observed during training, as illustrated in Fig.~\ref{fig:gap}. Existing 4K video methods typically address this limitation through additional high-resolution training or fine-tuning~\cite{ultravideo,hu2026ultragen}, which incurs substantial computational and data costs. To the best of our knowledge, no prior method enables video diffusion models trained at low resolutions to generate substantially higher-resolution, high-fidelity videos at inference time without additional supervision.

\section{Preliminaries}
\label{sec:preliminary}

\noindent \textbf{Self-Attention} is one of the core operations in DiTs, enabling each visual token to aggregate information from all other tokens in the latent representation. Given a latent representation at diffusion step $t$, denoted as $\phi(z_t)$, self-attention computes pairwise interactions among the latent tokens through learned query, key, and value projections. Attention can be formulated as,
\begin{equation}
\mathrm{Attn}(\phi(z_t))
=
\mathrm{Softmax}\!\left(
\frac{QK^\top}{\sqrt{d}}
\right)V,
\end{equation}
\noindent where
\(
Q=\ell_Q\phi(z_t),
K=\ell_K\phi(z_t),
\)
and
\(
V=\ell_V\phi(z_t)
\)
are the query, key, and value projections of the latent tokens, respectively, and $d$ denotes the projection dimensionality. The attention weights determine how strongly each token aggregates information from every other token, allowing the model to capture long-range spatial and temporal dependencies throughout the video. Unlike convolutional neural networks with limited receptive fields, self-attention establishes global interactions between all visual tokens.

\begin{figure*}[t]
\centering
\includegraphics[width=0.99\linewidth]{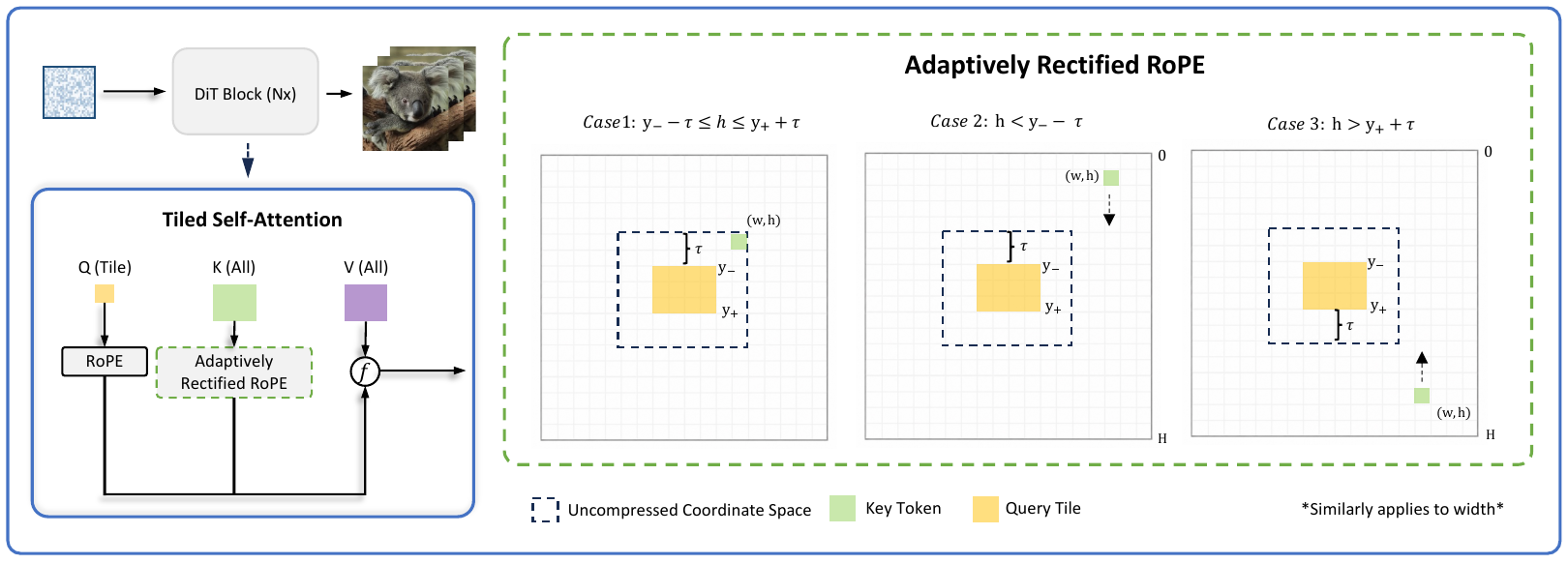}
\caption{
\textbf{Overall framework of CineScale.} 
Generating videos at resolutions substantially beyond the training regime exposes positional encodings to out-of-distribution spatial coordinates. To address this, we introduce \emph{Adaptively Rectified RoPE}, which partitions queries into spatial tiles and constructs a tile-conditioned coordinate mapping for global keys. Keys within a local neighborhood of each query tile are encoded regularly, while more distant keys are monotonically compressed toward the tile and into the pretrained positional range. This preserves fine-grained local geometry while enabling local queries to attend consistently to global context during high-resolution inference.
}
\label{fig:framework}
\end{figure*}
\vspace{1.0em}
\noindent \textbf{Rotary Positional Embedding.} Without positional information, self-attention is permutation equivariant and therefore cannot distinguish different spatial-temporal arrangements of the same token features. Modern diffusion transformers, including Wan~\cite{wan2025}, address this limitation using Rotary Positional Embeddings (RoPE)~\cite{su2024roformer}. RoPE partitions each query and key into two-dimensional subspaces and rotates each pair according to the token's position. For position $n$, the $j$-th feature pair is transformed as
\begin{equation}
\mathcal{R}_{n,j}
=
\begin{bmatrix}
\cos(n\theta_j) & -\sin(n\theta_j) \\
\sin(n\theta_j) & \cos(n\theta_j)
\end{bmatrix},
\qquad
\theta_j=\beta^{-2j/d},
\end{equation}
where $\beta$ is the frequency base, $d$ is the rotary embedding dimension, and $j\in\{0,\ldots,d/2-1\}$. Applying these rotations to queries and keys causes their inner product to depend on the relative displacement between their positions:
\begin{equation}
\bigl(\mathcal{R}_{m}q_m\bigr)^\top
\bigl(\mathcal{R}_{n}k_n\bigr)
=
q_m^\top\mathcal{R}_{n-m}k_n.
\end{equation}
For video tokens, Wan applies this mechanism separately across the temporal, vertical, and horizontal coordinates. 
At spatial resolutions (e.g. 4K) substantially beyond those used during training (e.g. 720p), the spatial coordinates and relative offsets become out of distribution. Although the RoPE frequencies remain fixed, evaluating them at unseen offsets produces unfamiliar rotation angles, distorting learned spatial relationships and degrading structural coherence. Existing context-extension methods such as YaRN~\cite{peng2023yarn} rescale or interpolate RoPE frequencies to accommodate longer sequences. However, such global rescaling also alters the positional geometry of nearby tokens and was developed primarily for one-dimensional language sequences. In our experiments, directly applying this strategy to video diffusion does not enable high-fidelity video generation. We therefore introduce Adaptively Rectified RoPE. For each query tile, our method preserves the original relative coordinates of nearby keys while progressively compressing the relative offsets of more distant keys into the range encountered during training. This preserves local positional geometry while enabling attention across substantially higher-resolution video latents without producing out-of-distribution positional offsets.

% \begin{figure*}[t]
% \centering
% \includegraphics[width=0.99\linewidth]{Teaser2.pdf}
% \caption{
% \re{\textbf{Full Self-Attention vs Selective Self-Attention Ablation.} 
% We compare the effects of using selective self-attention and full self-attention. We find that using selective self-attention generally yields sharper results. Zoomed-in regions highlight sharper architectural structures, clearer human figures, and richer textures while preserving the global scene composition. 
% }}
% \label{fig:Teaser2}
% \end{figure*}

\section{CineScale}
\label{sec:methodology}
In this section, we describe the implementation of Adaptively Rectified RoPE.
\subsection{Tiled Self-Attention.} 

\noindent We partition the query tokens into \(M\) non-overlapping tiles
\(\{\mathcal{T}_m\}_{m=1}^{M}\), where each tile spans the corresponding spatial region across all video frames. Let \(\mathcal{I}_m\) denote the indices of the query tokens assigned to tile \(\mathcal{T}_m\). The queries for tile \(m\) are
\begin{equation}
\mathbf{Q}_m=\mathbf{Q}[\mathcal{I}_m,:].
\end{equation}
Each query tile attends to the complete set of global keys and values
\begin{equation}
\mathbf{O}_m
=
\operatorname{Softmax}\!\left(
\frac{
\widetilde{\mathbf{Q}}_m
\left(\widetilde{\mathbf{K}}_{\mathrm{all}}^{(m)}\right)^{\top}
}{
\sqrt{d_h}
}
\right)
\mathbf{V}_{\mathrm{all}}
\end{equation}
\noindent where \(\widetilde{\mathbf{Q}}_m\) contains the positionally encoded queries within tile \(m\), while \(\widetilde{\mathbf{K}}_{\mathrm{all}}^{(m)}\) and \(\mathbf{V}_{\mathrm{all}}\) contain all key and value tokens in the video, respectively. With standard RoPE, computing self-attention separately for each query tile and concatenating the outputs mathematically yields exactly the same result as conventional global self-attention. We instead apply \emph{Adaptively Rectified RoPE} to the keys, which assigns them tile-dependent spatial coordinates. The superscript \(m\) indicates that the positional encoding of the keys is conditioned on tile \(m\). Finally, the resulting tile outputs are scattered back to their original query locations:
\begin{equation}
\mathbf{O}[\mathcal{I}_m,:]=\mathbf{O}_m.
\end{equation}

\begin{figure*}[!t]
\centering
\includegraphics[width=0.99\linewidth]{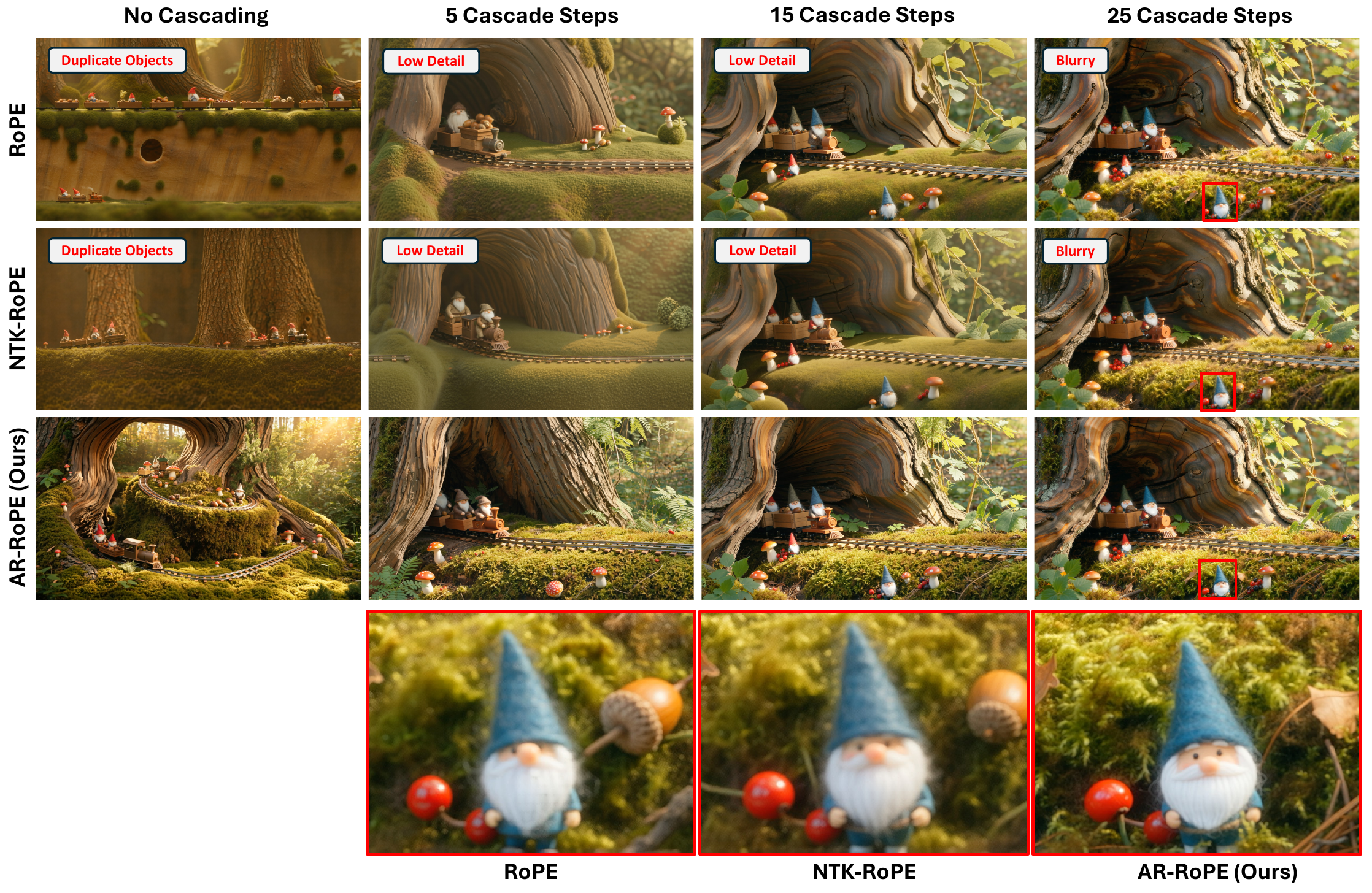}
\caption{\textbf{Ablation of RoPE variants at 4K resolution.} Standard RoPE and NTK-RoPE produce structural artifacts under large spatial extrapolation and yields over-smoothed details and jittering. Adaptively Rectified RoPE preserves local positional geometry while rectifying distant
offsets, resulting in sharper details and more coherent global structure.
}
\label{fig:RoPE Ablation}
\end{figure*}

\subsection{Adaptively Rectified RoPE}

Generating videos at resolutions far beyond those seen during training is problematic. At higher resolutions, the same normalized spatial displacement spans many more latent tokens, causing standard RoPE to encode relative offsets outside the model's training distribution. These unseen offsets can distort learned spatial relationships and undermine global structural coherence. We therefore introduce \emph{Adaptively Rectified RoPE}. Adaptively Rectified RoPE spatially encodes each key relative to each query that attends to it. Specifically, it preserves the original coordinates of keys near each query tile while compressing more distant key coordinates towards the query. This preserves precise local geometry while constraining long-range relative offsets to the range supported by the pretrained model, enabling stable long-range semantic interactions at higher resolutions.

\begin{figure*}[!t]
\centering
\includegraphics[width=0.99\linewidth]{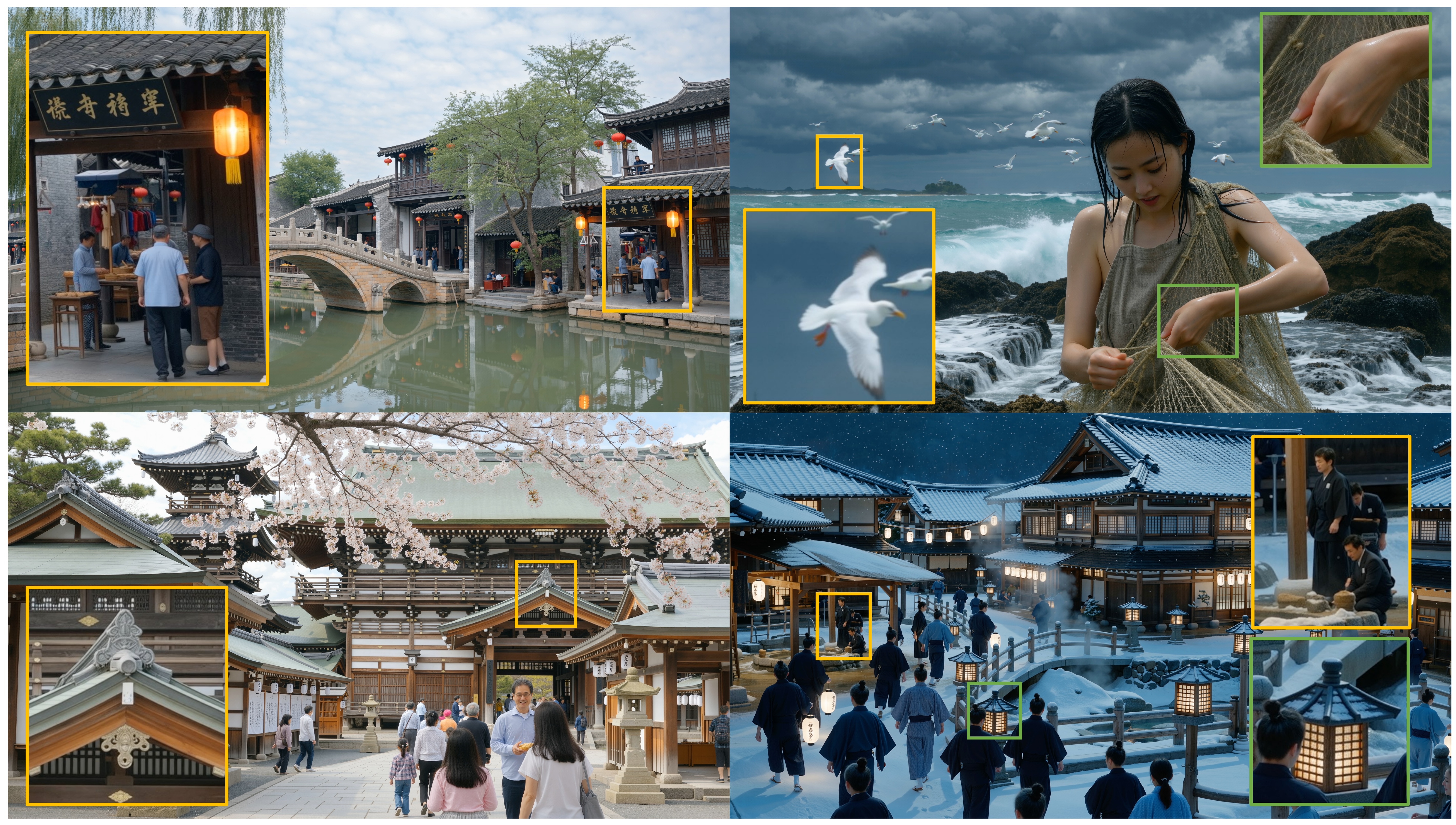}
\caption{
\textbf{Qualitative Results (4K Resolution)} The enlarged regions demonstrate CineScale's ability to synthesize sharp, high-frequency textures across diverse scenes, including intricate fabric embroidery, fine clothing patterns, architectural ornamentation, distant people and birds, and delicate object boundaries. These details remain visually coherent with the surrounding scene, demonstrating high-fidelity 4K generation beyond merely increasing output dimensions.
}
\label{fig:teaser_2}
\end{figure*}

We rectify the spatial coordinates of each key relative to each query tile. Consider the vertical axis \(y\in[0,L_y-1]\) and a query tile \(m\) spanning \([\ell_y^{(m)},u_y^{(m)}]\). Let \(R_y\) denote the maximum relative offset between a query and key, chosen according to the maximum offset supported by the pretrained model. We preserve the original positional geometry within the query tile, whose extent is \(D_y^{(m)}=u_y^{(m)}-\ell_y^{(m)}\). This consumes \(D_y^{(m)}\) of the supported relative positional range, leaving an external positional budget of \(B_y^{(m)}=R_y-D_y^{(m)}\). Given a local preservation threshold \(\tau_y\), we retain the original coordinates within
\begin{equation}
[\kappa_{y,-}^{(m)},\kappa_{y,+}^{(m)}]
=
[\ell_y^{(m)}-\delta_{y,-}^{(m)},\,
u_y^{(m)}+\delta_{y,+}^{(m)}],
\end{equation}
where
\begin{equation}
\begin{aligned}
\delta_{y,-}^{(m)}
&=\min(\tau_y,\ell_y^{(m)},B_y^{(m)}),\\
\delta_{y,+}^{(m)}
&=\min(\tau_y,L_y-1-u_y^{(m)},B_y^{(m)}).
\end{aligned}
\end{equation}
Coordinates outside this interval are linearly compressed toward the tile, with endpoints

\begin{equation}
\begin{aligned}
\gamma_{y,-}^{(m)}
&=\max\!\left(0,u_y^{(m)}-R_y\right),\\
\gamma_{y,+}^{(m)}
&=\min\!\left(L_y-1,\ell_y^{(m)}+R_y\right).
\end{aligned}
\end{equation}

\noindent For a global key at coordinate \(p_y\), the rectified coordinate is:

\begin{equation}
\rho_y^{(m)}(p_y)=
\begin{cases}
\gamma_{y,-}^{(m)}+\lambda_{y,-}^{(m)}p_y,
& p_y<\kappa_{y,-}^{(m)},\\[3pt]

p_y,
& p_y\in[\kappa_{y,-}^{(m)},\kappa_{y,+}^{(m)}],\\[3pt]

\begin{aligned}
&\kappa_{y,+}^{(m)}
+\lambda_{y,+}^{(m)}\\[-2pt]
&\qquad\cdot
\bigl(p_y-\kappa_{y,+}^{(m)}\bigr)
\end{aligned}
&
p_y>\kappa_{y,+}^{(m)}.
\end{cases}
\end{equation}

\begin{equation}
\lambda_{y,-}^{(m)}
=
\frac{\kappa_{y,-}^{(m)}-\gamma_{y,-}^{(m)}}
{\kappa_{y,-}^{(m)}},
\qquad
\lambda_{y,+}^{(m)}
=
\frac{\gamma_{y,+}^{(m)}-\kappa_{y,+}^{(m)}}
{L_y-1-\kappa_{y,+}^{(m)}}.
\end{equation}

\noindent The horizontal mapping \(\rho_x^{(m)}\) is defined analogously. Importantly, rectification modifies only the positional coordinates assigned to the keys; the query/key features and query coordinates remain unchanged. Thus, for a key at \((t,y,x)\),
\begin{align}
\widetilde{\mathbf Q}_m
&=\operatorname{RoPE}(t,y,x)\mathbf Q_m,\\
\widetilde{\mathbf K}_{\mathrm{all}}^{(m)}
&=\operatorname{RoPE}\!\left(
t,\rho_y^{(m)}(y),\rho_x^{(m)}(x)
\right)\mathbf K_{\mathrm{all}}.
\end{align}
This preserves local positional geometry while compressing long-range offsets into the pretrained RoPE range, enabling stable global interactions at 4K resolution.

\begin{figure*}[!t]
\centering
\includegraphics[width=0.99\linewidth]{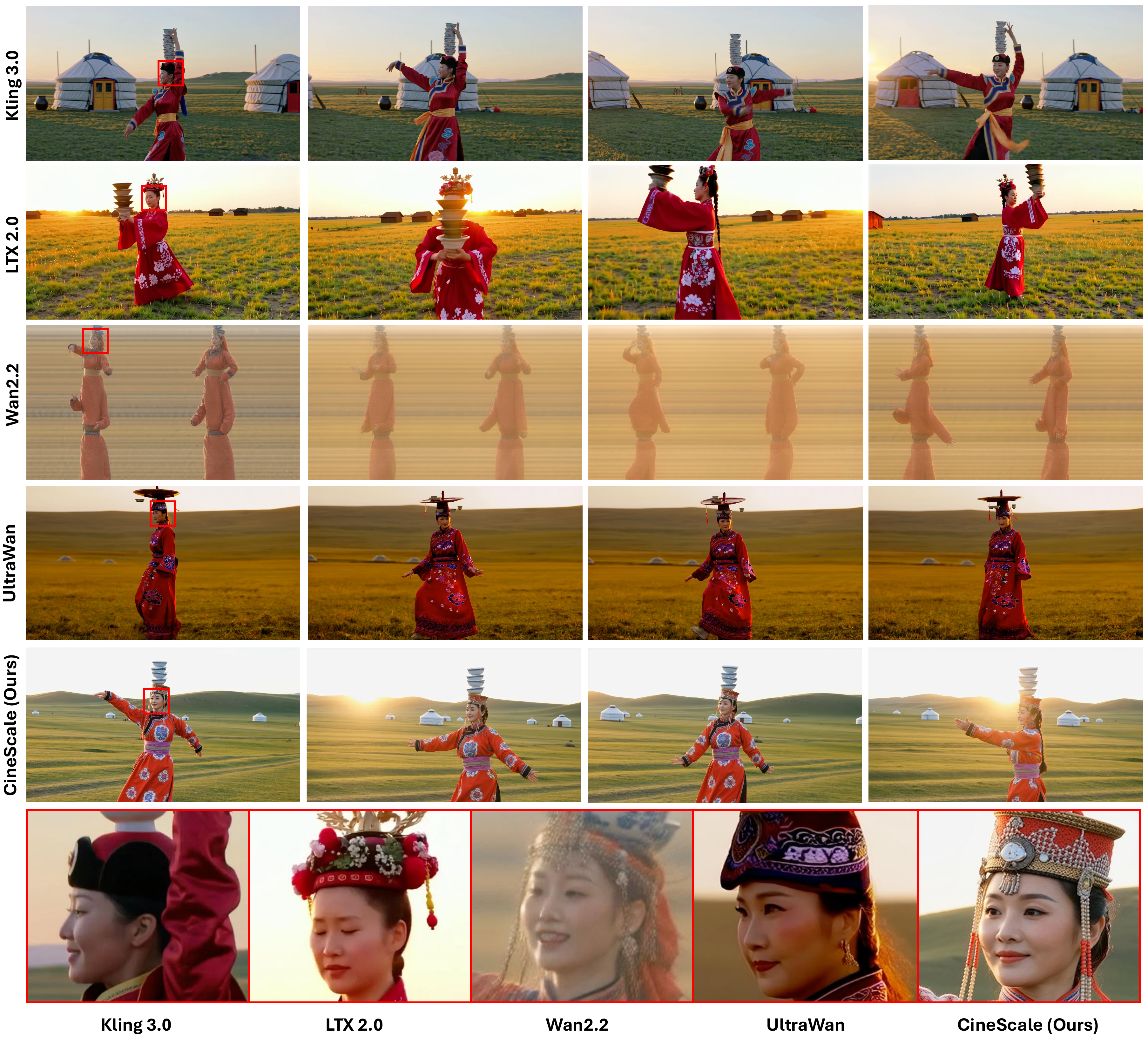}
\caption{
\textbf{Qualitative Results (4K Resolution)} 
 Comparison of high-resolution video generation across recent video generation models. Even state-of-the-art models (e.g., Kling 3.0, LTX 2.0) trained explicitly on 4K data generate blurry results at this resolution. Wan2.2 suffers from noticeably degraded global coherence, with severe blurring and duplicated structures. In contrast, CineScale preserves far more high-frequency details such as facial features, headdresses, embroidery patterns, hair strands, and textures on the robe while maintaining better global consistency. The resulting visual quality is competitive with state-of-the-art models that are explicitly trained on 4K videos. 
}
\label{fig:4k_comparison}
\end{figure*}

\subsection{Self-Cascade Upscaling}

Similar to most frameworks~\cite{du2024demofusion,guo2024make}, we first generate a video at the model’s native training resolution. The generated video is then upsampled through bilinear interpolation in the latent space. We then perturb the upsampled latent with Gaussian noise at an intermediate diffusion timestep and denoise it to recover fine details while preserving the coarse structure established by the low-resolution generation. During this refinement stage, we apply Adaptively Rectified RoPE to support self-attention beyond the model's native spatial range. This can significantly reduce runtime without sacrificing visual fidelity, as demonstrated in Fig.~\ref{fig:RoPE Ablation}.

% \begin{figure*}[t]
% %
% \centering
% \includegraphics[width=0.99\linewidth]{2k_comparison.pdf}
% \caption{
% \re{\textbf{Qualitative Results (1440p Resolution)} 
%  Comparison of high-resolution video generation across recent video generation models. Although CineScale is a lightweight inference-time method built on top of Wan2.2 without retraining the backbone, it substantially enhances fine-grained details, producing sharper facial features, hair, clothing textures, and scene geometry than the original Wan2.2. The resulting visual quality is competitive with recent state-of-the-art models that are explicitly trained on high-resolution videos (Veo3.1 and Kling 3.0 use 1080p). Best viewed ZOOMED-IN. 
% }}
% \label{fig:2k_comparison}
% \end{figure*}

\section{Experiments}
\label{sec:experiments}

\subsection{Experimental Setup.}
We use Wan2.1-T2V-1.3B for quantitative evaluation to enable direct comparisons with existing methods built on the same backbone, and Wan2.2-T2V-A14B for qualitative evaluation to demonstrate scalability to larger models, without updating any model parameters. All experiments use the default configuration and use 8 A800 GPUs to run inference. We first generate a low-resolution video, then bilinearly upsample its latent to the target resolution. The upsampled latent is then perturbed at a late diffusion timestep and refined for some denoising steps using Adaptively Rectified RoPE, preserving the global structure while recovering high-frequency details. We use 25 refinement steps for Wan2.2 and 35 steps for Wan2.1. We set the maximum relative key-query offsets according to the models' training resolution, which is a maximum of 720p for both backbones, thereby keeping the rectified positional offsets within the pretrained spatial range. For AR-RoPE, Wan2.2 uses non-overlapping \(32\times18\) query tiles in latent-token space with \(\tau_x=16\) and \(\tau_y=9\), while Wan2.1 uses \(21\times12\) query tiles with \(\tau_x=16\) and \(\tau_y=9\). For the best results, we recommend generating at up to approximately 3× the base model’s native resolution. For example, if a model natively generates at 480p, target up to roughly 1440p (2K); if it natively generates at 720p, target up to roughly 2160p (4K).

\begin{table*}[!t]
\centering
\caption{\textbf{VBench comparison across different target resolutions.}
The best, second-best, and third-best results are marked in
\textbf{bold}, \underline{underline}, and \textit{italics}, respectively.}
\label{tab:vbench}
\begin{tabular*}{\textwidth}{@{\extracolsep{\fill}}l|cccccc@{}}
\toprule
\textbf{Method} &
\textbf{SC}$\uparrow$ &
\textbf{BC}$\uparrow$ &
\textbf{TF}$\uparrow$ &
\textbf{AQ}$\uparrow$ &
\textbf{IQ}$\uparrow$ &
\textbf{Average}$\uparrow$ \\
\midrule
\multicolumn{7}{l}{\textit{Tuning-Free}} \\
Wan2.1-720p         & 0.9570 & 0.9605 & 0.9845 & 0.5646 & 0.6828 & 0.8299 \\
Wan2.1-1K           & 0.9540 & 0.9645 & 0.9898 & 0.4989 & 0.5826 & 0.7980 \\
Wan2.1-4K           & 0.9470 & 0.9760 & \underline{0.9950} & 0.2880 & 0.3740 & 0.7160 \\
CineScale-2K (Ours) & \textit{0.9734} & \underline{0.9777} & 0.9795 & \textbf{0.6488} & \underline{0.7156} & \textbf{0.8581} \\
\midrule
\multicolumn{7}{l}{\textit{Tuning-Based}} \\
UltraWan-1K         & 0.9586 & 0.9661 & 0.9853 & 0.5686 & 0.6966 & 0.8350 \\
UltraWan-4K         & 0.9581 & 0.9611 & 0.9771 & 0.5769 & \textit{0.7144} & 0.8375 \\
UltraGen-1080P      & \underline{0.9771} & \underline{0.9777} & \textbf{0.9961} & 0.5819 & \textbf{0.7350} & \underline{0.8536} \\
UltraGen-4K         & \textbf{0.9854} & \textbf{0.9894} & \textit{0.9933} & 0.5787 & 0.6832 & \textit{0.8460} \\
LUVE-2K             & 0.9583 & 0.9676 & 0.9818 & \underline{0.5978} & 0.7115 & 0.8434 \\
LUVE-4K             & 0.9536 & 0.9646 & 0.9809 & \textit{0.5891} & 0.7133 & 0.8403 \\
\bottomrule
\end{tabular*}
\end{table*}

\subsection{Evaluation Metrics.}

Since high-resolution inference methods aim to preserve the content and motion of the original generation while enhancing spatial detail, we evaluate subject consistency (SC), background consistency (BC), temporal flickering (TF), aesthetic quality (AQ), and imaging quality (IQ) using VBench~\cite{vbench}. We randomly sample 100 augmented prompts from the VBench~\cite{vbench,huang2024vbench++} dataset and release the complete subset for reproducibility. Following the evaluation setting adopted by prior high-resolution video generation methods, we evaluate 29-frame videos. We compare CineScale against recent high-resolution video generation methods, including direct Wan~\cite{wan2025} inference, LUVE~\cite{zhao2026luve}, UltraGen baselines~\cite{hu2026ultragen}, and UltraWan~\cite{ultravideo}. Where model weights or evaluation code are unavailable, we report published results from~\cite{hu2026ultragen}.

\section{Results}

Our ablation in Fig.~\ref{fig:RoPE Ablation} reveals distinct failure modes under high-resolution extrapolation. Standard RoPE extrapolates to spatial offsets far beyond the pretrained range, causing duplicated objects, structural inconsistency, local blurriness, and temporal jitter. NTK-RoPE~\cite{peng2023yarn} reduces long-range extrapolation by rescaling the RoPE frequencies, but this also compresses local positional distinctions, resulting in over-smoothed details and similar local jitter. Increasing the number of cascade steps partially mitigates these artifacts, but cannot recover the lost high-frequency structure. In contrast, Adaptively Rectified RoPE preserves local positional geometry while compressing only distant offsets into the pretrained range, yielding more stable structure, reduced jitter, and sharper fine-grained details.

Table~\ref{tab:vbench} shows that CineScale substantially improves perceptual quality at high resolution, especially in aesthetic and imaging quality. Subject consistency, background consistency, and motion
smoothness remain comparable to those of the underlying base model,
because these properties are primarily determined by the base model's
generative capabilities rather than explicitly optimized by CineScale. Our method instead focuses on increasing spatial fidelity and recovering
fine-grained visual details while preserving the temporal behavior and
semantic structure produced by the base model. 

Qualitatively, CineScale is even competitive with recent state-of-the-art video generation models that are explicitly optimized or trained for higher-resolution generation. The advantage becomes more pronounced at 4K resolution. As shown in Fig.~\ref{fig:4k_comparison}, direct Wan2.2 inference degrades severely, producing duplicated band-like human structures and poor global coherence. This failure is consistent with the positional distribution shift caused by applying RoPE to spatial coordinates far beyond the model's training regime. Other high-resolution models such as Kling 3.0 and LTX can preserve reasonable global structure, but their outputs remain relatively blurry in zoomed-in regions. CineScale instead maintains coherent scene layout while generating richer high-frequency details, including clearer facial structure, sharper clothing patterns, more detailed accessories, and cleaner object boundaries. Overall, the quantitative and qualitative results indicate that CineScale improves high-resolution visual fidelity without sacrificing temporal smoothness or semantic consistency, enabling pretrained video diffusion models to generate substantially sharper 4K videos even when they are trained on videos up to 720p.

\section{Conclusion}
\label{sec:conclusion}

In this paper, we presented CineScale, a tuning-free inference framework for high-resolution video generation with DiT-based diffusion models. We showed that out-of-distribution spatial positional offsets are a major source of degradation when pretrained video diffusion models are scaled substantially beyond their native training resolution. To address these challenges, we introduced Adaptively Rectified RoPE, which rectifies positional coordinates to remain consistent with the distribution encountered during pretraining. Extensive experiments demonstrate that CineScale enables stable video generation at resolutions substantially beyond the training resolution of the backbone model without requiring any parameter updates or additional supervision. Our approach consistently improves perceptual quality, local detail, and visual fidelity while maintaining temporal coherence. Although CineScale substantially improves high-resolution inference, ultra-high-resolution video generation remains computationally demanding due to the increasing cost of processing long spatiotemporal token sequences. Future work includes developing more efficient sparse attention mechanisms, adaptive token routing strategies, and scalable positional representations to further improve both efficiency and generation quality at even higher resolutions.

\section{Acknowledgements}

This research is supported by the National Research Foundation, Singapore under its AI Singapore Programme (AISG Award No: AISG2-PhD-2022-01-035T), the Ministry of Education, Singapore, under its MOE AcRF Tier 2 (MOE-T2EP20221-0012, MOE-T2EP20223-0002), and Netflix Eyeline Studios.

\section{Ethics Statement}
Our paradigm, based on the pre-trained video diffusion model, assists the model in generating high-resolution videos. As a result, regulators only need to oversee the original video generation model to ensure adherence to ethical standards. 

\section{Data Availability Statement}
Models used in this paper are all publicly available online. CineScale's source code  repository is publicly available at \href{https://github.com/Eyeline-Labs/CineScale}{https://github.com/Eyeline-Labs/CineScale}.

% % BibTeX users please use one of
% \bibliography{ref}   % name your BibTeX data base

{
    \small
    \bibliographystyle{ieeenat_fullname}
    \bibliography{main}
}

\clearpage
\setcounter{page}{1}
\maketitlesupplementary

\section{Preliminaries}

\noindent \textbf{Video Diffusion Models}~\cite{wan2025} typically operate in a latent space by encoding a video sample
$\mathbf{x}$,
\begin{equation}
\mathbf{z}_0=\mathcal{E}(\mathbf{x}).
\end{equation}
\noindent A noisy latent at time $t\in[0,1]$ is constructed by linear interpolation between a Gaussian sample $\boldsymbol{\epsilon}\sim\mathcal{N}(0,\mathbf{I})$ and the clean latent, which can be formulated as,
\begin{equation}
\mathbf{z}_t=(1-t)\boldsymbol{\epsilon}+t\mathbf{z}_0.
\end{equation}
\noindent In the case of flow matching, a neural network $v_\theta(\mathbf{z}_t,t)$ is trained to predict a continuous vector field that transports the noisy latent $\mathbf{z}_t$ to the target latent representation $\mathbf{z}_0$.  During inference, generation starts from pure Gaussian noise and numerically integrates the learned ordinary differential equation,
\begin{equation}
\frac{d\mathbf{z}_t}{dt}=v_\theta(\mathbf{z}_t,t),
\end{equation}
from $t=0$ to $t=1$, producing the final latent representation $\mathbf{z}_0$, which is decoded back into the pixel space.

\section{VBench Evaluation Prompts}

We use the following randomly sampled augmented VBench prompts for quantitative evaluation. To facilitate reproducibility and enable consistent future comparisons, we release the complete set of evaluation prompts below.

\begin{CJK*}{UTF8}{gbsn}

\begingroup
\small
\begin{enumerate}[label=\arabic*.,leftmargin=2.2em,labelsep=0.5em,itemsep=0.25em,parsep=0pt,topsep=0.25em]

\item 3D model of a 1800s Victorian house, standing tall\ldots{}

\item A憨态可掬的大熊猫坐在巴黎一家咖啡馆里，优雅地啜饮着一杯拿铁咖啡。熊猫皮毛黑白分明，眼睛圆溜溜的，嘴角带\ldots{}

\item A憨态可掬的大熊猫站在冲浪板上，脚下的波浪在日落的余晖下闪烁着金色光芒。大熊猫身披淡蓝色的冲浪服，毛发闪\ldots{}

\item A blue ceramic bowl sits on a rustic wooden table in\ldots{}

\item A casual street scene captured in a vibrant urban\ldots{}

\item A close-up shot of a traditional Japanese\ldots{}

\item A confused panda student wearing a classic academic\ldots{}

\item A couple in elegant formal evening wear, the man in\ldots{}

\item A couple in formal evening wear, both dressed in\ldots{}

\item A dramatic lightning strike illuminates the iconic\ldots{}

\item A futuristic robot dancer with metallic blue skin\ldots{}

\item A gentle cow standing gracefully in a lush green\ldots{}

\item A joyful fuzzy panda playing a small electric guitar\ldots{}

\item A laptop and a remote control sitting on a wooden\ldots{}

\item A leisurely boat sailing along the Seine River,\ldots{}

\item A leisurely boat sailing along the picturesque Seine\ldots{}

\item A leisurely boat sailing along the serene Seine\ldots{}

\item A person in a casual summer attire, wearing\ldots{}

\item A person is engrossed in a chess game, their fingers\ldots{}

\item A person is roller skating down a winding city\ldots{}

\item A person motorcycling through a bustling city street\ldots{}

\item A person reading a book in a cozy library setting.\ldots{}

\item A picturesque coastal beach during the spring season\ldots{}

\item A potted plant standing tall in a corner of a modern\ldots{}

\item A simple yet elegant mug filled with steaming hot\ldots{}

\item A sleek black motorcycle weaving skillfully through\ldots{}

\item A sleek black sports car with a racing stripe on its\ldots{}

\item A tranquil tableau of a bathroom, featuring a serene\ldots{}

\item A tranquil tableau of a beautiful, handcrafted\ldots{}

\item A tranquil tableau of an exquisite mahogany dining\ldots{}

\item A tranquil tableau of the Stonehenge presented\ldots{}

\item A vibrant person engaging in energetic aerobics at a\ldots{}

\item A vintage school bus slowly turning a corner on a\ldots{}

\item A vintage steam locomotive speeds down the tracks at\ldots{}

\item A vintage wooden suitcase with polished wood grain\ldots{}

\item A whimsical animated short film, featuring a gentle\ldots{}

\item A whimsical animated short film, a playful kitten\ldots{}

\item A whimsical scene captured in a vibrant summer park,\ldots{}

\item A whimsical scene captured in a vibrant summer\ldots{}

\item A yellow bird perched gracefully on a branch of a\ldots{}

\item Action-packed scene from a sports documentary, a\ldots{}

\item Amusement park themed cinematic video, vibrant neon\ldots{}

\item An aerial perspective video showcasing an airplane\ldots{}

\item An animated short film, an African elephant standing\ldots{}

\item Animated style, a majestic great white shark\ldots{}

\item CG animated digital art, Gwen Stacy reading a\ldots{}

\item CG animation digital art, a majestic bird perched on\ldots{}

\item CG animation digital art, a playful panda with a\ldots{}

\item CG fantasy digital art, Yoda, the wise green alien,\ldots{}

\item CG fantasy landscape digital art, a vast snowy\ldots{}

\item CG game concept digital art, a tranquil tableau\ldots{}

\item CG game concept digital art, a human gently petting\ldots{}

\item CG game concept digital art, a detailed close-up\ldots{}

\item CG game concept digital art, a muscular man with\ldots{}

\item CG game concept digital art, a person sitting at a\ldots{}

\item CG game concept digital art, a majestic giraffe\ldots{}

\item CG game concept digital art, a large wooden baseball\ldots{}

\item CG game concept digital art, a large tennis racket\ldots{}

\item CG game concept digital art, a large dining table\ldots{}

\item CG game concept digital art, a vibrant pink bird\ldots{}

\item CG game concept digital art, a large black bowl with\ldots{}

\item CG game concept digital art, a majestic great white shark swimming\ldots{}

\item CG game concept digital art, a majestic great white shark\ldots{}

\item CG game concept digital art, an astronaut in a sleek, silver spacesuit floating weightlessly in the vast expanse of space. The astronaut has a calm, focused expression with their helmet tilted slightly. They are\ldots{}

\item CG game concept digital art, an astronaut in a sleek, silver spacesuit floating weightlessly in the vast expanse of space. The astronaut has a calm, focused expression with their helmet tilted slightly. They are wearing\ldots{}

\item CG game concept digital art, a cozy bathroom with a\ldots{}

\item CG game concept digital art, a majestic lighthouse\ldots{}

\item CG game concept digital art, a marsh landscape with\ldots{}

\item CG kitchen concept art, a bunch of broccoli sitting\ldots{}

\item CG space concept digital art, an astronaut in a\ldots{}

\item In a modern kitchen setting, a sleek black hair\ldots{}

\item In a serene still frame, a tranquil pond was fringed\ldots{}

\item In a still frame, a ripe pear sits on a rustic\ldots{}

\item In a still frame, the Temple of Hephaestus,\ldots{}

\item Macro slo-mo, slow-motion closeup shot of freshly\ldots{}

\item Snow-covered rocky mountains with their peaks and\ldots{}

\item Van Gogh-inspired astronaut flying in outer space,\ldots{}

\item Van Gogh-inspired snowy landscape, towering rocky\ldots{}

\item Winter sports scene, skis and a snowboard lying next\ldots{}

\item A beautiful coastal beach in spring, waves gently\ldots{}

\item A bowl filled with colorful fresh fruits, such as\ldots{}

\item A minimalist scene capturing the essence of everyday\ldots{}

\item A modern urban scene captured in front view,\ldots{}

\item A person is barbequing outdoors on a warm summer\ldots{}

\item A person swimming in the vast expanse of the ocean,\ldots{}

\item A picturesque coastal beach during the spring season,\ldots{}

\item A picturesque scene captured in fresh film\ldots{}

\item A remote control device sits elegantly on the right\ldots{}

\item A sleek modern microwave oven sits atop a stainless\ldots{}

\item A tranquil tableau of a cozy restaurant, nestled\ldots{}

\item A vintage steam train gliding gracefully on the\ldots{}

\item A whimsical animated short film, featuring a playful\ldots{}

\item A whimsical scene captured in watercolor animation,\ldots{}

\item An orange perfectly placed atop a carrot, in a\ldots{}

\item CG food concept digital art, a vibrant hot dog\ldots{}

\item CG game concept digital art, a snowboarder with a\ldots{}

\item CG game concept digital art, a human gently petting\ldots{}

\item CG game concept digital art, skis placed on the\ldots{}

\item CG kitchen appliance concept art, a sleek black\ldots{}

\item Winter mountain scene, two pairs of skis placed side\ldots{}

\end{enumerate}
\endgroup

\end{CJK*}

\section{Applications}

\begin{figure}[t]
\centering
\includegraphics[width=0.99\linewidth]{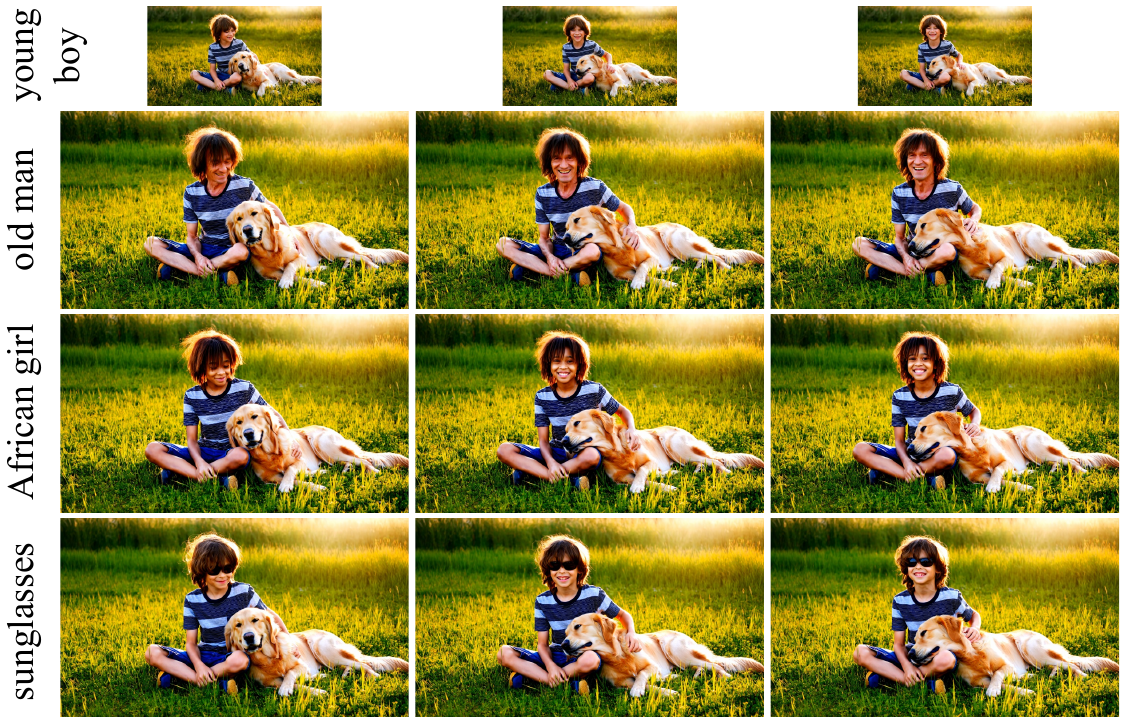}
\caption{\textbf{Local semantic editing for video generation.} CineScale supports efficient editing by allowing users to preview results at low resolution while modifying high-resolution local semantics via prompts.}
\label{fig:t2v_edit}
\end{figure}

\noindent\textbf{Local Semantic Editing for Video Generation.} The self-cascade upscaling paradigm in CineScale naturally supports an efficient and user-friendly editing workflow. Based on low-resolution previews, users can adjust prompts during high-resolution denoising to perform localized semantic edits. As shown in Figure~\ref{fig:t2v_edit}, starting with a low-resolution image depicting a boy and a dog cuddling on the grass, users can refine the scene through prompt editing once the initial composition is satisfactory. For example, the boy can be transformed into an elderly man, replaced with an African girl by altering gender and ethnicity, or modified with accessories like sunglasses. These changes are accurately reflected in the final high-resolution video.

\begin{figure*}[t]
\centering
\includegraphics[width=0.99\linewidth]{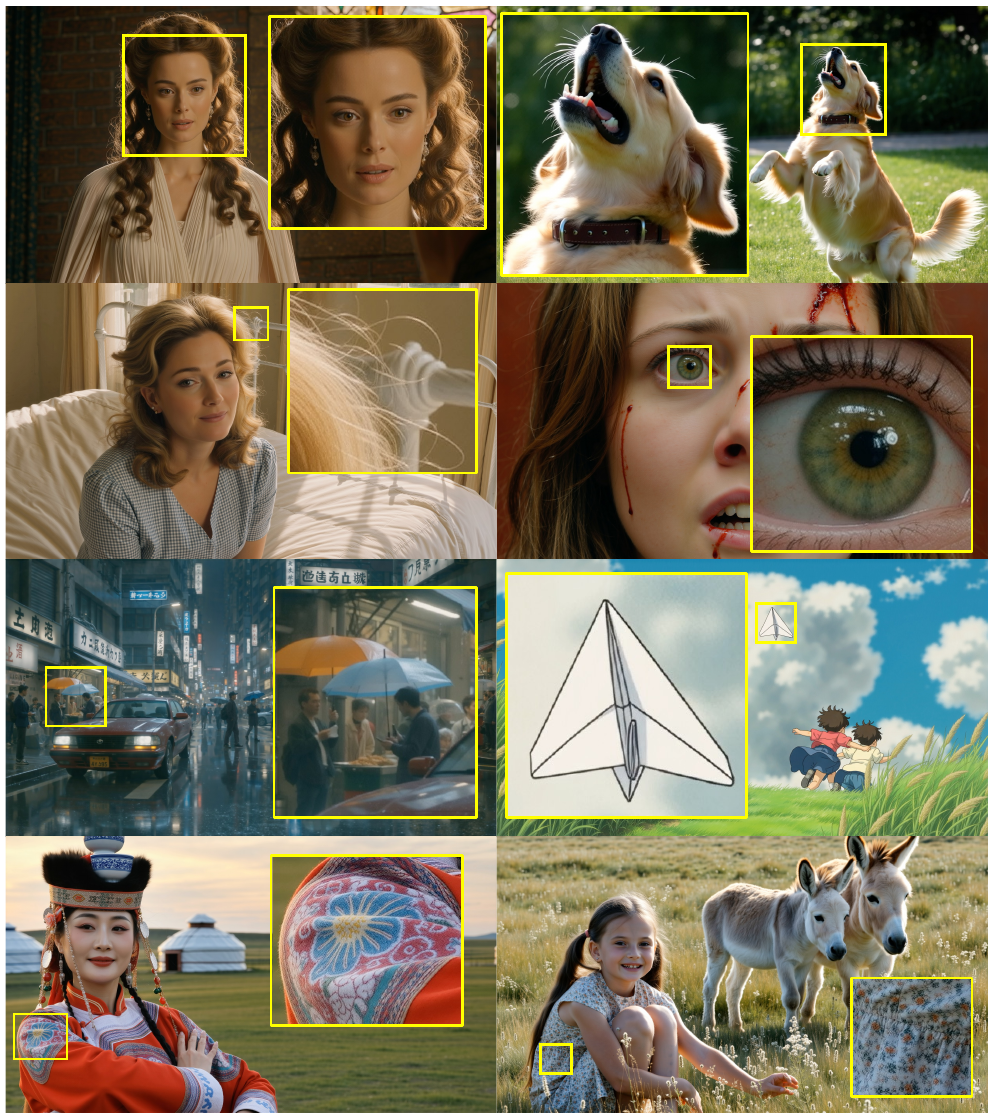}
\caption{
\textbf{Additional Qualitative Results (4K Resolution)} The enlarged regions demonstrate CineScale's ability to synthesize sharp, high-frequency textures across diverse scenes, including intricate fabric embroidery, fine clothing patterns, architectural ornamentation, distant people and birds, and delicate object boundaries. These details remain visually coherent with the surrounding scene, demonstrating high-fidelity 4K generation beyond merely increasing output dimensions.
}
\label{fig:teaser_3}
\end{figure*}

\section{Limitations}
\label{sec:limitations}

\noindent\textbf{Base Model Bottleneck.} Our method improves the spatial and image fidelity of a pretrained video diffusion model but does not substantially improve the model's motion prior, or temporal reasoning. As a result, failure cases inherited from the base model, such as incorrect object interactions, unnatural motion, or prompt misalignment, may still persist.

\noindent \textbf{Long Inference Time.} Although Adaptively Rectified RoPE introduces no trainable parameters, 4K generation requires substantially longer token sequences, leading to high memory and runtime costs. Our implementation therefore relies on tiled query processing and multi-GPU inference, limiting accessibility on consumer hardware.

\noindent \textbf{DiT-Based Exclusivity.} CineScale is designed for DiT-based video diffusion models using RoPE-style positional encodings. Its direct applicability to models with different positional encoding schemes, sparse attention mechanisms, or non-transformer architectures remains unexplored.

% The key issue is that NTK-aware RoPE changes the positional geometry not only for long-range interactions, but also for local interactions that the model already learned correctly during training.NTK-RoPE modifies the RoPE frequency spectrum to accommodate larger coordinate ranges. This effectively makes positional phase changes vary more slowly along some frequency dimensions. Consequently, nearby spatial positions can become less distinguishable in positional space than they were during training.

\end{document}